\documentclass{llncs}
\usepackage[utf8]{inputenc}
\usepackage{amsmath}
\usepackage{amssymb}	
\usepackage{amsfonts}	
\usepackage{hyperref}
\usepackage{graphicx}
\usepackage{xcolor}
\usepackage{notations}

\title{Fast and Scalable Optimal Transport \\ for Brain Tractograms}
\titlerunning{OT for brain tractograms}
\author{Jean Feydy\inst{1,2*} \and Pierre Roussillon\inst{3}\thanks{contributed equally to this work} \and Alain Trouvé\inst{1} \and Pietro Gori\inst{3}}
\authorrunning{J. Feydy et al.}
\institute{CMLA, ENS Paris-Saclay, France \and DMA, École Normale Supérieure, Paris, France \and LTCI, Télécom ParisTech, Institut Mines Télécom, Paris, France}

\begin{document}
\maketitle              
\begin{abstract}
We present a new multiscale algorithm for solving regularized Optimal Transport problems on the GPU, with a linear memory footprint. Relying on Sinkhorn divergences which are convex, smooth and positive definite loss functions, this method enables the computation of transport plans between millions of points in a matter of minutes. We show the effectiveness of this approach on brain tractograms modeled either as bundles of fibers or as track density maps. We use the resulting smooth assignments to perform label transfer for atlas-based segmentation of fiber tractograms. The parameters -- \emph{blur} and \emph{reach} -- of our method are meaningful, defining the minimum and maximum distance at which two fibers are compared with each other. They can be set according to anatomical knowledge. Furthermore, we also propose to estimate a probabilistic atlas of a population of track density maps as a Wasserstein barycenter. Our CUDA implementation is endowed with a user-friendly PyTorch interface, freely available on the \texttt{PyPi} repository (\texttt{pip install geomloss}) and
at \url{www.kernel-operations.io/geomloss}.
\end{abstract}

\section{Introduction}

\noindent\textbf{Optimal Transport.}
Matching unlabeled distributions is a fundamental problem
in mathematics.
Traced back to the work of Monge in the 1780's,
Optimal Transport (OT) theory is all about 
finding \emph{transport plans} 
that minimize a ground cost metric under marginal constraints,
which ensure a full covering of the input measures
-- see \eqref{eq:linear_primal} and \cite{peyre2019computational} for a modern overview.
Understood as a canonical way of lifting distances
from a set of \emph{points} to a space of \emph{distributions},
the OT problem has appealing geometric properties:
it is robust to a large class of deformations, ensuring a perfect
retrieval of global translations \emph{and} small perturbations.
Unfortunately though, standard combinatorial solvers
for this problem scale in $O(n^3)$ with the number of samples. Until recently, OT has thus been mostly restricted
to applications in economy and operational research,
with at most a thousand points per distribution.

\noindent\textbf{Fast computational methods.}
In the last forty years, a considerable amount of work
has been devoted to approximate numerical schemes
that allow users to \emph{trade time for accuracy}.
The two most popular iterative solvers for large-scale Optimal Transport,
the \emph{auction} and the 
\emph{Sinkhorn} (or \emph{IPFP}, \emph{SoftAssign}) algorithms, 
can both be understood as \emph{coordinate ascent} updates
on a relaxed dual problem~\cite{kosowsky1994invisible}:
assuming a tolerance $\varepsilon$ on the final result,
they allow us to solve the transportation problem
in time $O(n^2 / \varepsilon)$. They met widespread diffusion in the computer vision
literature at the turn of the XXI$^{\text{st}}$ century \cite{chui2003new}.

\noindent\textbf{Recent progress.}
As of 2019, two lines of work strive to take advantage of the growth of (parallel) computing power to speed up large-scale Optimal Transport.
The first track, centered around \emph{multiscale strategies},
leverages the inner structure of the data with
a coarse-to-fine scheme~\cite{schmitzer2016stabilized}.
It was recently showcased to the medical imaging
community~\cite{gerber2018exploratory} and
provides solvers that generally scale in $O(n\log(n / \varepsilon))$
on the CPU, assuming an efficient 
octree-like decomposition of the input data. On the other hand, a second line of works puts a strong focus
on entropic regularization and highly-parallel, GPU implementations of 
the Sinkhorn algorithm~\cite{cuturi2013sinkhorn,peyre2019computational}.
Up to a clever \emph{debiasing} of the regularized 
OT problem~\cite{ramdas2017wasserstein},
authors in the Machine Learning community
were able to provide the first strong
\emph{theoretical guarantees} for an approximation of Optimal Transport,
from principled extensions for the \emph{unbalanced} 
setting \cite{chizat2018interpolating}
to proofs of positivity, definiteness and
convexity~\cite{feydy2019interpolating,sejourne2019unbalanced}.

\noindent\textbf{Our contribution.}
This paper is about merging together these two bodies of work,
as we provide the \emph{first GPU implementation of a 
multiscale OT solver}.
Detailed in section~\ref{sec:algorithm}, our multiscale Sinkhorn algorithm
is a \emph{symmetric} coarse-to-fine scheme that takes
advantage of key ideas developed in the past five years.
Leveraging the routines of the
\texttt{KeOps} library~\cite{keops}, our \texttt{PyTorch} implementation
re-defines the state-of-the-art for discrete OT,
allowing users to handle high-dimensional feature spaces in a matter of minutes.
It is freely available on the \texttt{PyPi} repository 
(\texttt{pip install geomloss}) and at \url{www.kernel-operations.io/geomloss}.


\noindent\textbf{White matter segmentation.}
From diffusion MR images, 
probabilistic tractography algorithms can reconstruct the architecture of the human brain white matter as bundles of 3D polylines \cite{zhang_anatomically_2018}, called fibers, or as track density maps \cite{wassermann_unsupervised_2010}. In clinical neurology and neurosurgery, 
a task of interest is the segmentation of the white matter into anatomically relevant tracts.
This can be carried out by: 1- manually delineating Regions of Interest (ROIs), which is tedious and time-consuming;  2- directly modeling the anatomical definitions \cite{delmonte_white_2019}; 3- using learning strategies \cite{wasserthal_tractseg_2018}; or 4- transferring labels of an atlas via non-linear deformations and clustering algorithms \cite{sharmin_white_2018,garyfallidis_recognition_2018}. This last class of methods usually depends on several hyperparameters (e.g. number of clusters, kernel size). 
In this paper, similarly to \cite{sharmin_white_2018}, we propose to use Optimal Transport for transferring the labels of a fiber atlas to a subject tractogram. Leveraging the proposed efficient and multi-resolution implementation, we are able to directly segment a whole brain tractogram without any pre-processing or clustering. The entire algorithm only depends on two meaningful hyperparameters, the \emph{blur} and \emph{reach} scales, that define the minimum and maximum distances at which two fibers or densities are compared with each other. On top of label transfer, we also propose to estimate a probabilistic atlas of a population of track density maps as a Wasserstein barycenter where each map describes, for every voxel in the space, the probability that a specific track (e.g. IFOF) passes through.


\section{Methods}
\label{sec:algorithm}

We now give a detailed exposition of our \emph{symmetric},
\emph{unbiased}, \emph{multiscale} Sinkhorn algorithm for discrete OT.
We will focus on \emph{vector data}
and assume that our discrete, positive measures
$\alpha$ and $\beta$ are encoded as weighted sums of Dirac masses:
$\alpha = \textstyle\sum_{i=1}^\N \alpha_i\,\delta_{x_i}$ 
and $\beta = \textstyle\sum_{j=1}^\M \beta_j\,\delta_{y_j}$
with weights $\alpha_i, \beta_j \geqslant 0$ and
samples' locations $x_i, y_j \in \Xx = \R^\D$.
We endow our feature space $\Xx = \R^\D$
with a cost function $\C(x,y)=\tfrac{1}{p}\|x-y\|^p$,
with $p \in [1,2]$ ($p=2$ in this paper), and recall the standard 
Monge-Kantorovitch transportation problem:
\begin{align}
\OT(\alpha,\beta)~&=~
\min_{\pi \in \R^{\N\times \M}_{\geqslant 0}}
~\textstyle\sum_{i,j} \pi_{i,j}\,\C(x_i,y_j) \quad \text{s.t.}\quad
(\pi \,\ones)_i = 
\alpha_i, ~ (\pi^{\intercal} \ones)_j = \beta_j.
\label{eq:linear_primal} 
\end{align}
In most applications, the feature space $\Xx$ is the ambient space $\R^3$ 
endowed with the standard Euclidean metric. 
This is the case, for instance, when using \emph{track density maps}
where $\alpha_i$ and $\beta_j$ are the probabilities associated to the voxel 
locations $x_i$ and $y_j$, respectively. 
Meanwhile, when using fiber tractograms, a usual strategy is to 
resample each fiber to the same number of points $\P$. 
In this case, the feature space $\Xx$ becomes $\R^{\P\times 3}$
and
$x_i$, $y_j$ are the $\N$ and $\M$ fibers that constitute 
the source and target tractograms
with uniform weights $\alpha_i = \tfrac{1}{\N}$ and
$\beta_j = \tfrac{1}{\M}$,
respectively.
Distances can be computed using the standard Euclidean $L^2$ norm 
-- normalized by $1/\sqrt{\P}$ -- 
and we alleviate the problem of \emph{fiber orientation} 
by augmenting our tractograms with the mirror flips of all fibers.
This corresponds to the simplest of all encodings 
for unoriented curves. 

\noindent\textbf{Robust, regularized Optimal Transport.}
Following \cite{chizat2018interpolating,peyre2019computational},
we consider a generalization of \eqref{eq:linear_primal} where $\alpha$ and $\beta$ don't have
the same total mass or may contain \emph{outliers}: this is typically the case when working with fiber bundles. 
Paving the way for efficient computations, the relaxed OT problem reads:
\begin{align}
\OT_{\varepsilon,\rho}(\alpha,\beta)~&=~
\min_{\pi \in \R^{\N\times \M}_{\geqslant 0}}
~\langle \pi_{i,j}\,,\,\C(x_i,y_j) \rangle
~+~ \varepsilon\,\KL(\pi_{i,j},\alpha_i \otimes\beta_j) \label{eq:OTe_primal} \\[-.2cm]
&\qquad\qquad\qquad ~+~ \rho\,\KL((\pi\ones)_i, \alpha_i) 
~+~ \rho\,\KL((\pi^\intercal \ones)_j,\beta_j) \nonumber \\
&=~ \max_{f\in\R^N, g\in\R^M} ~
\rho\, \langle \alpha_i, 1 - e^{-f_i/\rho}\rangle
~+~
\rho \,\langle \beta_j, 1 - e^{-g_j/\rho}\rangle \label{eq:OTe_dual}\\[-.1cm]
~&\qquad\qquad\qquad +~
\varepsilon \,\langle \alpha_i\otimes \beta_j, 1 -
\exp \tfrac{1}{\varepsilon}\big[ f_i\oplus g_j - \C(x_i,y_j) \big] \rangle,\nonumber
\end{align}
where 
$\KL(a_i,b_i) = \sum a_i \log(\tfrac{a_i}{b_i}) - a_i + b_i$
denotes the (generalized) Kullback-Leibler divergence 
and $\varepsilon, \rho > 0$
are two \emph{positive} regularization parameters
homogeneous to the cost function $\C$.
The equality between the primal
and dual problems (\ref{eq:OTe_primal}--\ref{eq:OTe_dual}) 
-- \emph{strong duality} -- holds
through the Fenchel-Rockafellar theorem. With a \emph{linear} memory footprint,
optimal dual vectors $f$ and $g$ encode
a solution of the $\OT_{\varepsilon,\rho}$ problem given by:
\begin{align}
(\pi_{\varepsilon,\rho})_{i,j} ~=~ \alpha_i\beta_j~\exp \tfrac{1}{\varepsilon}\big[ f_i+ g_j - \C(x_i,y_j) \big] .\label{eq:primal_dual}
\end{align}
\noindent\textbf{Unbiased Sinkhorn divergences.}
Going further, we consider 
\begin{align}\hspace{-.3cm}
\S_{\varepsilon,\rho}(\alpha,\beta)
=
\OT_{\varepsilon,\rho}(\alpha,\beta)
-\tfrac{1}{2}\OT_{\varepsilon,\rho}(\alpha,\alpha)
-\tfrac{1}{2}\OT_{\varepsilon,\rho}(\beta,\beta)
+\tfrac{\varepsilon}{2}({\textstyle \sum} \alpha_i  
- {\textstyle \sum} \beta_j )^2,
\end{align}
the \emph{unbiased} Sinkhorn divergence
that was recently shown to define a \emph{positive}, \emph{definite}
and \emph{convex} loss function for measure-fitting applications
-- see \cite{feydy2019interpolating} for a proof
in the balanced setting, extended in
\cite{sejourne2019unbalanced} to the general case.

Generalizing ideas introduced by \cite{feydy2018global},
we can write $\S_{\varepsilon,\rho}$ in \emph{dual} form as
\begin{align}
\S_{\varepsilon,\rho}(\alpha,\beta)
~=~&
-~(\rho+\tfrac{\varepsilon}{2})\,\langle \alpha_i, e^{-b^{\beta\rightarrow\alpha}_i/\rho}
-
e^{-a^{\alpha\leftrightarrow\alpha}_i/\rho}
\rangle\\
&-~(\rho+\tfrac{\varepsilon}{2})\,\langle \beta_j, e^{-a^{\alpha\rightarrow\beta}_j/\rho} 
-
e^{-b^{\beta\leftrightarrow\beta}_j/\rho}
\rangle,\nonumber
\end{align}
where 
$(f_i,g_j) = (b^{\beta\rightarrow\alpha}_i, a^{\alpha\rightarrow\beta}_j)$ is a
solution of $\OT_{\varepsilon,\rho}(\alpha,\beta)$ and
$a^{\alpha\leftrightarrow\alpha}_i$,
$b^{\beta\leftrightarrow\beta}_j$
correspond to the unique solutions
of $\OT_{\varepsilon,\rho}(\alpha,\alpha)$
and $\OT_{\varepsilon,\rho}(\beta,\beta)$
on the diagonal of the space of dual pairs.
We can then derive the equations at optimality
for the three $\OT_{\varepsilon,\rho}$ problems
and write the gradients
$\partial_{\alpha_i} \S_{\varepsilon,\rho}$, 
$\partial_{x_i} \S_{\varepsilon,\rho}$,
$\partial_{\beta_j} \S_{\varepsilon,\rho}$, 
$\partial_{y_j} \S_{\varepsilon,\rho}$
as expressions of the \emph{four dual vectors}
$b^{\beta\rightarrow\alpha}_i, a^{\alpha\leftrightarrow\alpha}_i \in \R^\N$,
$a^{\alpha\rightarrow\beta}_j, b^{\beta\leftrightarrow\beta}_j \in \R^\M$  \cite{feydy2019interpolating,peyre2019computational}.

\noindent\textbf{Multiscale Sinkhorn algorithm.}
To estimate these dual vectors,  we propose a \emph{symmetrized} Sinkhorn loop
that retains the iterative structure of the
baseline SoftAssign algorithm, but replaces alternate updates
with \emph{averaged} iterations -- for the sake of symmetry -- and uses the $\varepsilon$-scaling 
heuristic of \cite{kosowsky1994invisible}:

\begin{algorithmic}[1]
\algrule[1pt]
\Statex \textbf{Parameters:}~Positive \textbf{blur}
and \textbf{reach} scales; 
\Statex \textbf{Multiscale heuristic:}~\textbf{diameter} $d = \max_{i,j} \|x_i-y_j\|$;
\textbf{scaling} $q = 0.9$.
\algrule
\State $a^{\alpha\leftrightarrow\alpha}_i,~
b^{\beta\leftrightarrow\beta}_j,~
a^{\alpha\rightarrow\beta}_j, ~
b^{\beta\rightarrow\alpha}_i
~\gets~ \mathbf{0}_{\R^\N}, ~\mathbf{0}_{\R^\M},~
\mathbf{0}_{\R^\M}, ~\mathbf{0}_{\R^\N}$
\vspace{.1cm}

\For{$\sigma$ \textbf{in} $[\, d, \,d\cdot q,\, d\cdot q^2,\, \cdots,\, \text{blur}\,]$} 
\Comment{$\lceil \log(d / \text{blur}) / \log(1/q) \rceil$ iterations}
\vspace{.1cm}

\State $\varepsilon,~ \rho,~ \lambda ~\gets~ \sigma^p,~ \text{reach}^p,~
1 / (1 + (\sigma/\text{reach})^p)$\vspace{.1cm}

\State $\tilde{a}^{\alpha\leftrightarrow\alpha}_i
\gets - \lambda\,\varepsilon 
\log \sum_{k=1}^\N 
\alpha_k~ \exp \tfrac{1}{\varepsilon}[ \,
a^{\alpha\leftrightarrow\alpha}_k
-
\C(x_k,x_i)\,] $

\State $\tilde{b}^{\beta\leftrightarrow\beta}_j\,
\gets - \lambda\,\varepsilon 
\log \sum_{k=1}^\M 
\beta_k~ \exp \tfrac{1}{\varepsilon}[ \,
b^{\beta\leftrightarrow\beta}_k\,
-
\C(y_k,y_j)\,] $

\State $\tilde{a}^{\alpha\rightarrow\beta}_j
\gets - \lambda\,\varepsilon 
\log \sum_{k=1}^\N 
\alpha_k~ \exp \tfrac{1}{\varepsilon}[ \,
b^{\beta\rightarrow\alpha}_k\,
-
\C(x_k,y_j)\,] $

\State $\tilde{b}^{\beta\rightarrow\alpha}_i\,
\gets - \lambda\,\varepsilon 
\log \sum_{k=1}^\M 
\beta_k~ \exp \tfrac{1}{\varepsilon}[ 
a^{\alpha\rightarrow\beta}_k\,
-
\C(y_k,x_i)\,] $
\vspace{.1cm}

\State $a^{\alpha\leftrightarrow\alpha}_i,~
b^{\beta\leftrightarrow\beta}_j
~\gets~ 
\tfrac{1}{2}( 
a^{\alpha\leftrightarrow\alpha}_i + 
\tilde{a}^{\alpha\leftrightarrow\alpha}_i),~~
\tfrac{1}{2}( 
b^{\beta\leftrightarrow\beta}_j + 
\tilde{b}^{\beta\leftrightarrow\beta}_j)
$

\State $
a^{\alpha\rightarrow\beta}_j,~ 
b^{\beta\rightarrow\alpha}_i
~\gets~ 
\tfrac{1}{2}( 
a^{\alpha\rightarrow\beta}_j + 
\tilde{a}^{\alpha\rightarrow\beta}_j),~~
\tfrac{1}{2}( 
b^{\beta\rightarrow\alpha}_i + 
\tilde{b}^{\beta\rightarrow\alpha}_i)
$
\vspace{.1cm}

\EndFor
\State \Return{$a^{\alpha\leftrightarrow\alpha}_i,~
b^{\beta\leftrightarrow\beta}_j,~
a^{\alpha\rightarrow\beta}_j, ~
b^{\beta\rightarrow\alpha}_i$}
\Comment{Optimal dual vectors.}
\algrule[1pt]

\end{algorithmic}

\noindent


\noindent \textbf{Coarse-to-fine strategy.}
To speed-up computations and tend towards the $O(n \log(n))$ complexity of multiscale methods, we use a coarse subsampling of the $x_i$'s and $y_j$'s in the first few iterations. Here, we use a simple K-means algorithm, but other strategies could be employed. We can then use the coarse dual vectors to \emph{prune out} useless computations at full resolution and thus implement the
\emph{kernel truncation trick} introduced by \cite{schmitzer2016stabilized}. 
Note that to perform these operations \emph{on the GPU}, 
our code heavily relies on the \emph{block-sparse}, online, 
generic reduction routines provided by the \texttt{KeOps} library \cite{keops}.
Our implementation provides a \texttt{x1,000} speed-up when compared
to simple \texttt{PyTorch} GPU implementations of the Sinkhorn
loop~\cite{cuturi2013sinkhorn},
while keeping a linear (instead of quadratic) memory footprint. Benchmarks may be found on our website.


\noindent\textbf{The \emph{blur} and \emph{reach} parameters.}
In practice, the only two parameters of our algorithm are the \emph{blur} and \emph{reach} scales, which specify the temperature $\varepsilon$ and the strength $\rho$ of the soft covering constraints. Intuitively, \emph{blur} is the resolution 
of the finest details that we try to capture, while \emph{reach} acts as an upper bound on the distance that points may travel to meet their targets -- instead of seeing them as \emph{outliers}. In our applications, they define the minimum and maximum distances at which two fibers or densities are compared with each other.

\noindent\textbf{Label transfer.}
Given a subject tractogram $\alpha$ and a segmented fiber atlas 
$\beta$ with $\L$ classes, 
we propose to transfer the labels
from the atlas to the subject with an OT matching.
Having computed optimal dual vectors $(f_i,g_j) = (b^{\beta\rightarrow\alpha}_i, a^{\alpha\rightarrow\beta}_j)$ solutions of $\OT_{\varepsilon,\rho}(\alpha,\beta)$, we encode the atlas labels as one-hot vectors $\ell_j$ in the probability simplex of $\R^\L$, and compute soft segmentation scores $\text{Lab}_i = \tfrac{1}{\alpha_i}(\pi_{\varepsilon,\rho}\ell)_i$ using the implicit encoding~\eqref{eq:primal_dual}:
\begin{align}
\text{Lab}_i = \textstyle\sum_{j=1}^{\M}
\beta_j\,\ell_j~
\exp \tfrac{1}{\varepsilon} [\,
f_i + g_j - \C(x_i,y_j)\,
]
~\in~\R^\L.
\end{align}
If a fiber $x_i$ is mapped by $\pi_{\varepsilon,\rho}$ to target fibers $y_j$
of the atlas $\beta$, 
the marginal
constraints ensure that $\text{Lab}_i$ is a \emph{barycentric}
combination of the corresponding labels $\ell_j$ on the probability simplex.
However, if $x_i$ is an \emph{outlier}, the $i$-th line
of $\pi_{\varepsilon,\rho}$ will be almost zero
and $\text{Lab}_i$ will be close to $\mathbf{0}_{\R^\L}$. In this way, we can detect and discard aberrant fibers. Please note that the OT plan is computed on the
augmented data, using both original and flipped fibers; only one version of each fiber will be matched to the atlas and thus kept for the transfer of labels.

\noindent\textbf{Track density atlas.}
Given $K$ \emph{track density maps} of a specific bundle encoded 
as normalized volumes $\beta_1$, \dots, $\beta_\K$, 
we also propose to estimate a probabilistic atlas 
as a \emph{Wasserstein barycenter} (or \emph{Fréchet mean})
by minimizing $\tfrac{1}{\K} \sum_{k=1}^\K
\S_{\varepsilon,\rho}(\alpha, \beta_k)$ with respect 
to the points $x_i\in\R^3$ of the atlas. 
This procedure relies on the computation of
$\S_{\varepsilon,\rho}(\alpha, \beta_k)$ 
with the squared Wasserstein-2 distance when 
$p=2$, \emph{reach} $=+\infty$ and
\emph{blur} is equal to the voxel size.

\begin{figure}[htb]
  \centering
      {\includegraphics[width=%
      \textwidth]{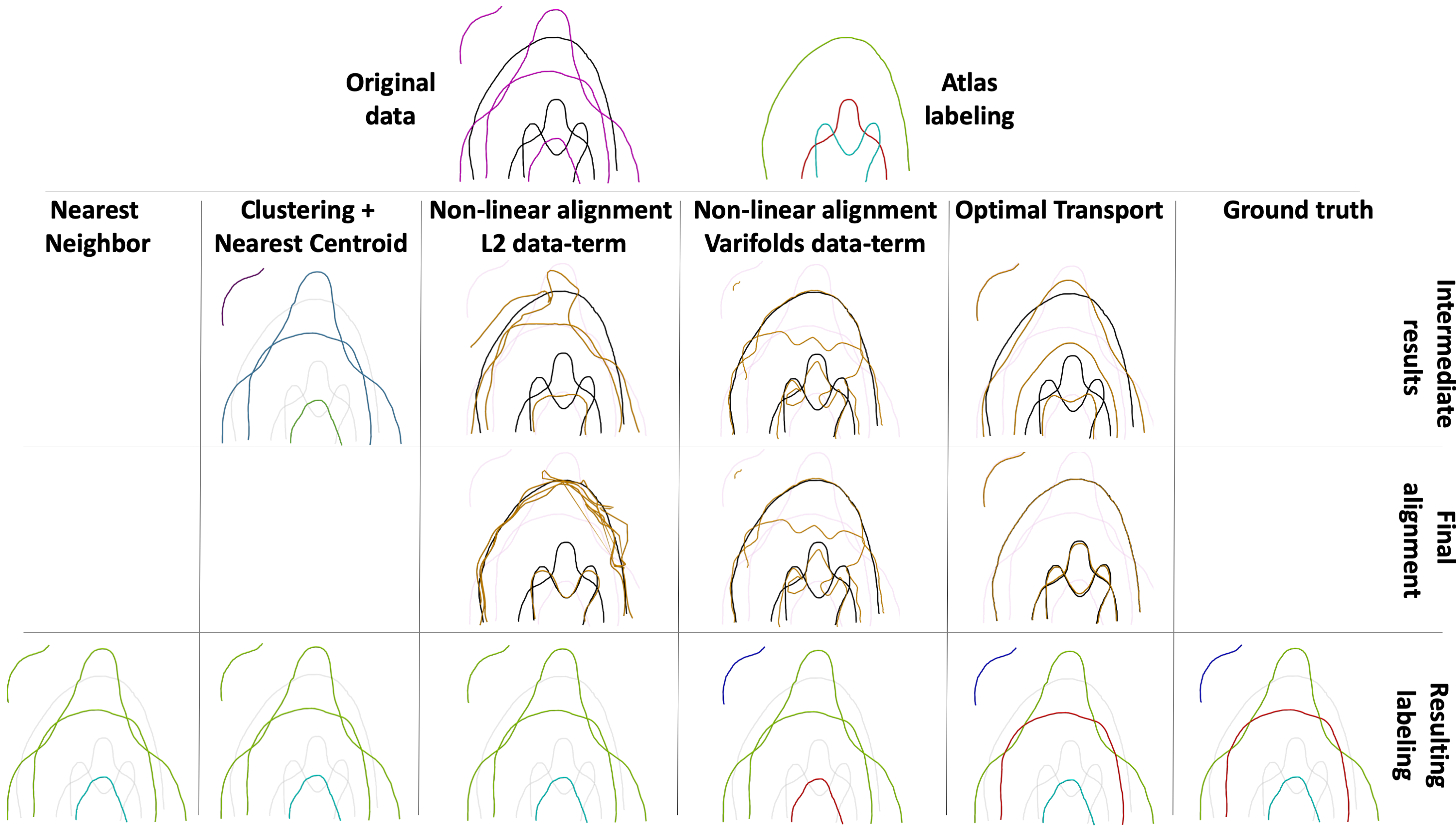}}       
  \caption{Label transfer between an atlas (black) and a subject (magenta) using different strategies. (1) Subject fibers are assigned the label of the closest atlas fiber. (2) Subject fibers are first clustered (K-means, K=3) and then each cluster is assigned the label of the closest atlas cluster. (3,4) Subject fibers are non-linearly aligned (i.e. diffeomorphism) to the atlas using a L2 data-term and a varifolds data-term. Labels are estimated from the resulting deformation. (5) Correspondences between subject and atlas fibers are estimated using the proposed algorithm for optimal transport. Second and third row show intermediate results (subject clustering and estimated alignement). In the last row, we present the transferred labels. Fibers detected as outliers are shown in dark blue.\vspace{-\baselineskip}} 
  \label{fig:toy_example}
\end{figure}

\begin{figure}[htbp]
  \centering
      {\includegraphics[width=11cm]{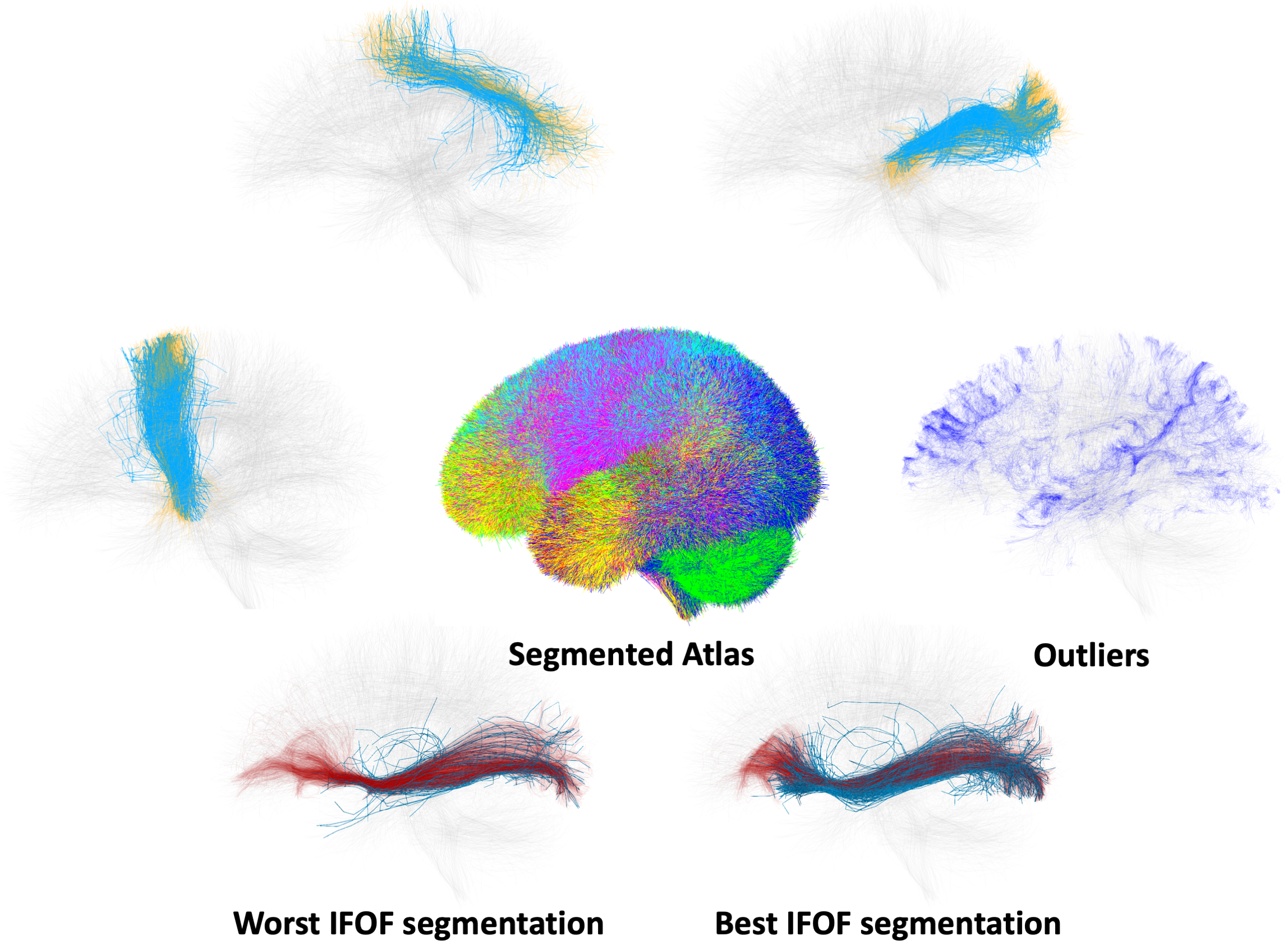}}   \caption{Label transfer between the segmented atlas (in the middle) and the subject tractograms. Top: some clusters of the atlas (in orange) with their respective segmentations (in light blue) of one random subject. Detected outliers are on the right (dark blue). Bottom: worst and best segmentation of the left IFOF, among the five tested subjects, compared to a manual segmentation.} 
  \label{fig:label-transfer}
\end{figure}

\begin{figure}[htbp]
  \centering
      {\includegraphics[width=11cm]{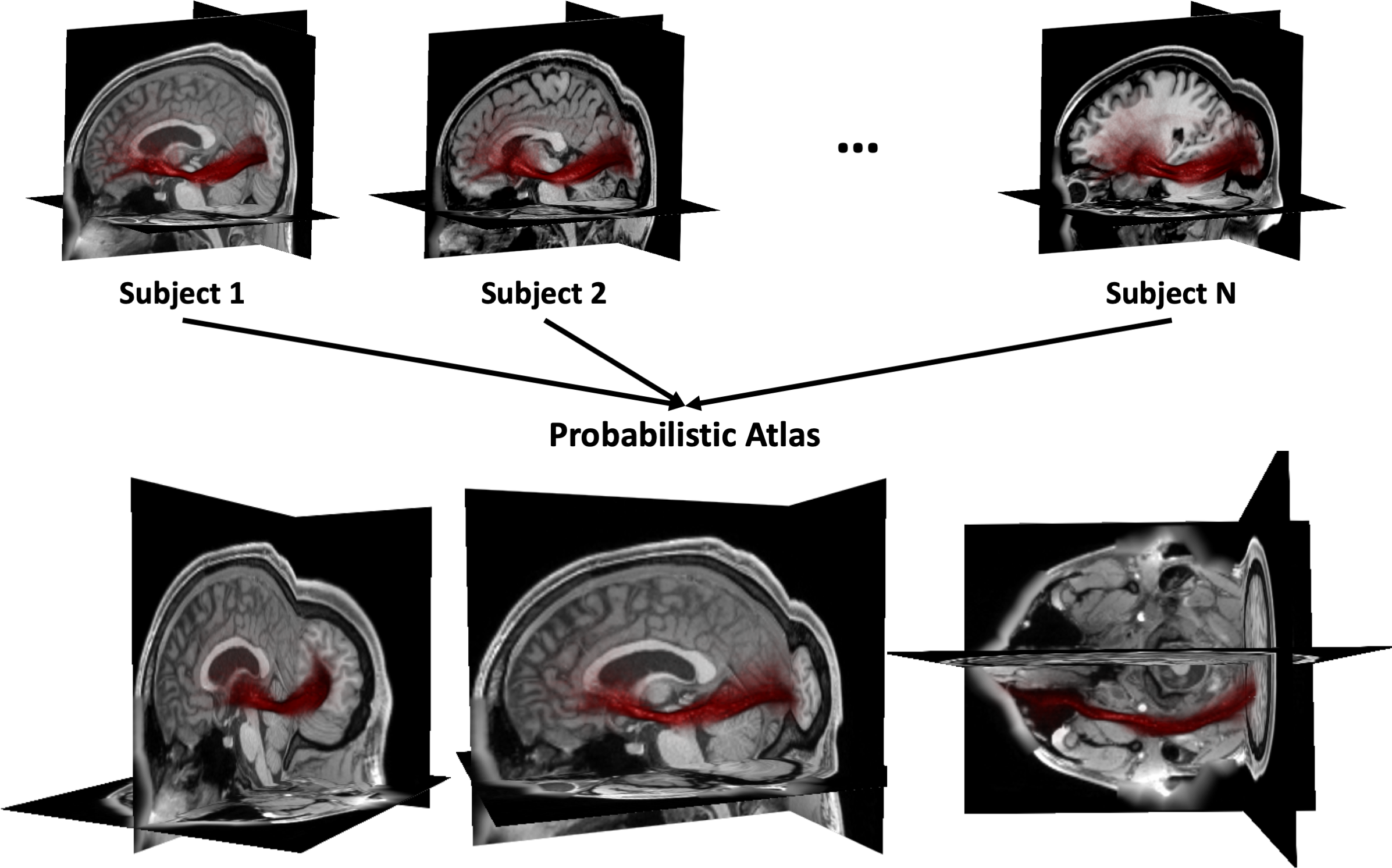}}       
  \caption{Probabilistic atlas of the left IFOF from 5 track density maps (in red). Top row: the densities of the five subjects, shown with their T1-wMRI.
  Bottom row: three views of the obtained atlas, alongside the T1-wMRI of one subject.} 
  \label{fig:barycenter}
\end{figure}

\section{Results and Discussion}

\noindent\textbf{Dataset.} Experiments below are based on 5 randomly selected healthy subjects of the \textit{HCP}\footnote{https://db.humanconnectome.org} dataset. Whole-brain tractograms of one million fibers are estimated with MRTrix3\footnote{http://www.mrtrix.org} using a probabilistic algorithm (iFOD2) and the Constrained Spherical Deconvolution model. Segmentations of the inferior fronto-occipital fasciculus (IFOF) are obtained as described in \cite{delmonte_white_2019}, and track density maps are computed using MRTrix3. We ran our scripts on an Intel Xeon E5-1620V4 with a GeForce GTX 1080.

\noindent\textbf{Toy example.} In Fig.~\ref{fig:toy_example}, we compare different existing strategies for transferring labels from a fiber atlas to a subject bundle with an U-shape. 
In this example, there is a bijection between the atlas and 
the subject bundle except for an outlier. 
The only method that estimates the correct correspondence and finds the outlier is the proposed one since, unlike the first two strategies, it takes into account the organization of the bundle. Note that for the second strategy, a different value of $K$ would not have changed the resulting labeling. The third and fourth strategies are based on diffeomorphic deformations and, regardless of the data-term, cannot correctly align the crossing fibers. 

\noindent\textbf{Label transfer.} In Fig.~\ref{fig:label-transfer}, we show some of the labels transferred from the atlas proposed in \cite{zhang_anatomically_2018} towards one test subject of the HCP dataset. The atlas is composed of 800 clusters, with $\sim$ 800K fibers. To speed up computations and improve interpretability, we separately analyze the two brain hemispheres and the inter-hemispheric fibers -- here, we only show results for the left hemisphere. To validate our results, we also compare the proposed segmentation of the left IFOF with a manual segmentation. Results indicate that the estimated labels are coherent with the clusters of the atlas and with the manual segmentation. The \emph{blur} and the \emph{reach} are empirically set to $2$mm and $20$mm respectively. The computation time of a transport plan for the left hemisphere is about 10 minutes (with $\sim$ 250K and $\sim$ 200K fibers sampled with $\P=20$ points for the subject and the atlas, respectively).

\noindent\textbf{Track density atlas.} Using 5 track density maps of the left IFOF, we estimate a probabilistic atlas shown in Fig.~\ref{fig:barycenter}. Each density map is a volume of $145\times 174 \times 145$ voxels of 1$\mathrm{mm}^3$ where approximately 
20,000 voxels have a probability (i.e. $\beta_j$) different from zero. The barycenter is initialized with the arithmetic average of the dataset densities and then up-sampled (factor of 6) to increase the resolution, resulting in $\sim$400,000 Dirac samples. The \emph{blur} is equal to the voxel size
and the \emph{reach} is set to $+\infty$. The computation time to estimate the Wasserstein barycenter is 9 seconds. 



\noindent\textbf{Conclusion.}
We presented an efficient, fast and scalable algorithm for solving the regularised (entropic) Optimal Transport problem
and showed its effectiveness in two applications. First, OT plans can be used to transport labels from a fiber atlas to a subject tractogram. This method takes into account the organization of the bundles -- unlike standard nearest neighbours or clustering algorithms --, detects outliers and is not hampered by fiber crossings. Second, we use Sinkhorn divergences
to estimate a \emph{geometric} average of track density maps. 
In future works, we plan to study more relevant features for 
modeling white-matter fibers, including for instance directional and FA information.



\paragraph{Acknowledgement} This research was partially supported by Labex DigiCosme (ANR) as part of the program Investissement d'Avenir Idex ParisSaclay.

\bibliographystyle{splncs04}
\bibliography{MICCAI2019}

\end{document}